\documentclass{article}

\usepackage{arxiv}
\usepackage[square,numbers]{natbib}
\usepackage[utf8]{inputenc} 
\usepackage[T1]{fontenc}    
\usepackage{hyperref}       
\usepackage{url}            
\usepackage{booktabs}       
\usepackage{amsfonts}       
\usepackage{nicefrac}       
\usepackage{microtype}      
\usepackage{lipsum}		
\usepackage{graphicx}
\usepackage{doi}
\usepackage{tabularx}
\usepackage{graphicx}
\usepackage{amsmath}
\usepackage{mdframed}
\usepackage{longtable}
\newcolumntype{P}[1]{>{\raggedright\arraybackslash}p{#1}}
\usepackage{comment}
\usepackage{csquotes} 
\usepackage{multirow}
\usepackage{array}
\usepackage{listings}
\usepackage{color,array}
\usepackage{subfig}
\usepackage{multicol}
\usepackage[noend]{algpseudocode}
\usepackage{algorithm}
\usepackage{float} 
\usepackage{booktabs} 
\usepackage{hyperref}
\usepackage{graphicx}
\usepackage{orcidlink}
\usepackage{comment}
\usepackage{authblk}

\title{Over the Edge of Chaos? Excess Complexity as a Roadblock to Artificial General Intelligence} 
\author[1]{Teo Susnjak\orcidlink{0000-0001-9416-1435}}
\author[2]{Timothy R. McIntosh\orcidlink{0000-0003-0836-4266}}
\author[3]{Andre L. C. Barczak\orcidlink{0000-0001-7648-285X}}
\author[1]{Napoleon H. Reyes\orcidlink{0000-0002-0683-436X}}
\author[1]{Tong Liu\orcidlink{0000-0003-3047-1148}}
\author[4]{Paul Watters\orcidlink{0000-0002-1399-7175}}
\author[5]{Malka N. Halgamuge\orcidlink{0000-0001-9994-3778}}

\affil[1]{School of Mathematical and Computational Sciences, Massey University, Auckland, New Zealand}
\affil[2]{Cyberoo Pty Ltd, Surrey Hills, NSW, Australia}
\affil[3]{Centre for Data Analytics, Bond University, Gold Coast, Australia}
\affil[4]{Cyberstronomy Pty Ltd, Ballarat, VIC, Australia}
\affil[5]{Department of Information Systems and Business Analytics, RMIT University, Melbourne, VIC, Australia}

\renewcommand{\shorttitle}{Excess Complexity as a Roadblock to Artificial General Intelligence}

\begin{document}

\maketitle
\begin{abstract}
In this study, we explored the progression trajectories of artificial intelligence (AI) systems through the lens of complexity theory. We challenged the conventional linear and exponential projections of AI advancement toward Artificial General Intelligence (AGI) underpinned by transformer-based architectures, and posited the existence of critical points, akin to phase transitions in complex systems, where AI performance might plateau or regress into instability upon exceeding a critical complexity threshold.  We employed agent-based modelling (ABM) to simulate hypothetical scenarios of AI systems' evolution under specific assumptions, using benchmark performance as a proxy for capability and complexity. Our simulations demonstrated how increasing the complexity of the AI system could exceed an upper criticality threshold, leading to unpredictable performance behaviours. Additionally, we developed a practical methodology for detecting these critical thresholds using simulation data and stochastic gradient descent to fine-tune detection thresholds. This research offers a novel perspective on AI advancement that has a particular relevance to Large Language Models (LLMs), emphasising the need for a tempered approach to extrapolating AI's growth potential and underscoring the importance of developing more robust and comprehensive AI performance benchmarks.
\end{abstract}

\keywords{
Artificial General Intelligence (AGI),
Large Language Models (LLMs),
Complexity Theory in AI,
Criticality in AI Systems,
Agent-Based Modelling (ABM),
AI System Evolution,
Phase Transitions in AI,
AI Instability and Criticality Detection,
AI Complexity Thresholds.
}

\section{Introduction}

In the current landscape of artificial intelligence (AI) research, especially with the recent advancements in Large Language Models (LLMs), there is a growing discourse about the future direction and ultimate potential of AI capabilities \cite{alfonseca2021superintelligence,burns2023weaktostrong, bostrom2024deep,Mitchell2024debates,ZHAO2023when,mclean2023risks,buttazzo2023rise}.  LLMs are breaking AI benchmarks at ever-increasing rates \cite{mialon2023gaia} with each new iteration and release.  The discussions in academic literature have started to shift towards a more optimistic outlook on the possibility of the eventual emergence of Artificial General Intelligence (AGI), where the hypothetical intelligence of a machine at least attains the capacity to understand, learn, and apply its knowledge in a broad range of domains much like the cognitive abilities of a human \cite{kurzweil2005singularity,alfonseca2021superintelligence,Mitchell2024debates,buttazzo2023rise,McIntosh2024game,mcintosh2023google}, or even exceed it in the form of \textit{superintelligence} \cite{bostrom2017superintelligence}. This positive sentiment has been fueled by the rapid advancements and demonstrable abilities of generative AI technologies designed to emulate human-like responses and creativity \cite{Mitchell2023how,Mitchell2023challenge}, developed primarily using transformer-based architectures \cite{vaswani2017attention}. Not only have large corporations at the forefront of developing these technologies started to take AGI more seriously, but much of popular culture has been captured by the idea of AGI inevitability, with many experienced and eminent researchers recently becoming supporters of this possibility \cite{burns2023weaktostrong,Mitchell2023how}.  Notably, leading AI-innovating organisations such as OpenAI, Meta, and DeepMind remain optimistic about the prospect of achieving AGI, despite the lack of consensus on its precise definition \cite{mclean2023risks} and the paths to its realisation, reflecting a landscape where the goalpost for AGI may continually shift for the foreseeable future \cite{Mitchell2024debates}.

However, humans have the tendency to anthropomorphise and attribute greater intelligence to AI technologies based on their linguistic output, where this can be mistaken for deep and genuine understanding \cite{Mitchell2023how}. When coupled with the systemic cognitive biases that humans possess when attempting to extrapolate exponential growth patterns \cite{gilovich2002heuristics} as seen in recent AI capabilities, it cannot be entirely surprising that the future trajectory of AI capabilities is increasingly directed towards AGI. Should these forecasts prove to be accurate, a reasonable expectation is that the AI agents' intelligence and reasoning capabilities should be accompanied by a commensurate increase in their internal complexity. 

Against this backdrop, complexity theory offers a framework for evaluating the optimism surrounding AI's growth trajectory. This theoretical perspective provides a lens for describing emergent behaviours and properties inherent in complex cognitive architectures \cite{ye2018cognitive}, and additionally, it lends the concept of \textit{criticality} that denotes a state which systems may realise as they accrue significant complexity. The principle of criticality in this context refers to a threshold beyond which a system's coherence and functional integrity are compromised, potentially leading to systemic instability or functional collapse. Complexity theory thus provides a mechanism for hypothetically exploring these phenomena because it uniquely accommodates the nonlinear interactions and feedback loops that are typical in AI development.
Therefore, if we are to accept the assumption that the current trends of AI advancement will continue, this study posits a hypothesis that contends with the narrative of inevitable and indefinite AI performance expansion. Our work suggests that AI systems, particularly those embodying current LLM architectures and likely those that immediately follow, may inherently possess a ceiling to their complexity and, by extension, their cognitive capabilities. This hypothesis draws parallels with biological systems, notably the human brain, which achieves remarkable complexity underpinned by sophisticated regulatory mechanisms that manage and mitigate the risks associated with excessive complexity and the speed at which the complexity is acquired \cite{plenz2021self}. The absence of analogous regulatory mechanisms in AI systems may predispose them to a state of excess criticality that is characterised by diminished coherence and ultimately, functionality.

\subsection*{Contributions}

The key contributions of this study have been as follows:

\begin{enumerate}
    \item This study investigated the theoretical foundations of criticality in complex AI systems and developed theoretical models and empirical analyses for simulating AI behaviour at and beyond critical thresholds.
    
    \item We proposed a novel framework for detecting the onset of excess criticality in real-world AI systems via benchmark performance metrics.
    
    \item Our research has hypothesised and provided evidence for inherent limits to the complexity and capabilities of AI systems given current ANN architectures.
    
    \item We presented a rich literature review and a balanced perspective on future AI development and trajectories.
\end{enumerate}

\section{Background}

\subsection{The case for achievability of AGI}

The anticipation of AGI becoming a reality is generally based on several technological, theoretical and philosophical grounds. Frequently it is justified based on substantive technological and theoretical AI advancements made over time, with notable recent accelerations \cite{buttazzo2023rise}. This includes significant progress afforded by deep learning ANN architectures producing systems routinely outperforming human abilities. These have ranged from AI systems that defeat world champions in chess and Go, as well as those outperforming humans in medical diagnoses, through to others possessing creative and generative abilities in music and art, indistinguishable from human-created works \cite{buttazzo2023rise}. Included also are advancements in unsupervised and reinforcement learning that enable AI systems to acquire knowledge more autonomously and refine their decision-making and planning in dynamic environments which mirrors key aspects of human cognitive adaptability \cite{wu2023daydreamer}. Researchers \cite{wu2023daydreamer} have shown how such systems can display compelling properties involving planning and predicting future outcomes with glimpses of being able to build and reason over real-world models. In the context of LLMs, some eminent scientists \cite{Mitchell2023challenge} contend that they also implicitly develop comprehensive world models representing knowledge, human experience, and subjective realities through the prediction of textual patterns in large-scale datasets, a claim which has been confirmed by other notable studies \cite{li2022emergent} indicating the emergence of world representations during the training processes. Recent works \cite{wu2024visualizationofthought} have indicated that LLMs can perform spatial reasoning by emulating the human capability to create mental images of unseen objects through a process known as "the Mind’s Eye" which can be elicited through prompting strategies. 
The demonstration of the Theory of Mind (ToM), previously considered exclusive to humans has recently \cite{kosinski2024evaluating} been suggested as an emergent behaviour within LLMs, where ToM is the cognitive ability to perceive and ascribe mental states to oneself and others. The study claimed that a ToM equivalent of a six-year-old child may have spontaneously emerged merely as a byproduct of LLMs' acquiring human language abilities.
The recent progress in generative AI's capabilities to encompass multimodal capabilities across text, audio, image and video channels has pointed towards a further milestone suggesting the incorporation of yet another fundamental trait of human intelligence \cite{ZHAO2023when}.
Finally, the exponential growth in computational power and the development of tailored AI-specific hardware such as GPUs and TPUs, has suggested that the infrastructure needed for AGI is now becoming available as well as the belief that AI systems will be able to achieve incremental autonomous improvement \cite{liu2023ai,ZHAO2023when}. For many, these collective advancements not only demonstrate AI's growing proficiency across a range of cognitive tasks \cite{mcintosh2024binadequacy} and their increasing potential to achieve even more, but they also serve as signposts and a confirmation of the plausibility of achieving AGI in the foreseeable future \cite{McIntosh2024gametheoretic}.

\subsection{Criticality in Neuroscience and AI}

Neuroscientific research has increasingly leaned on complexity theory and concepts to enhance our understanding of brain function.
The study of complexity in various scientific fields focuses on understanding how interactions among numerous elements in dynamic systems lead to different outcomes. In physics, chemistry, biology, and ecology, researchers investigate how the interactions of many components, such as atoms, biomolecules, cells, organisms, and populations, result in the emergence of organized structures and patterns \cite{mainzer2020natural}.
In neuroscience, the human brain routinely undergoes \textit{phase transitions} that shift neuronal activity from one pattern to another, such as transitions from wakefulness to sleep or from normal to hyperactive states observed in seizures. These transitions occur when criticality is reached as a result of internal or external stimuli. These state transitions are said to alter when a \textit{control parameter} such as neuronal excitation or inhibition cross a critical point\cite{o2022critical}. 
A prevalent hypothesis suggests that the brain functions optimally at the point of criticality that is defined as a boundary between order and disorder—frequently referred to as the \enquote{edge of chaos}\cite{bilder2014creative,o2022critical}. This critical state is considered ideal for cognitive functions, as it allows the brain to maintain stability while maximizing adaptability to new information. Operating at this edge, the brain can optimise its computational power by balancing between excessive regularity, which restricts adaptability, and excessive randomness, which can compromise coherence\cite{o2022critical}, highlighting the dual nature of complexity in producing both order and chaos \cite{mainzer2020natural}. 

Similar to neuroscience, ANNs are also held to operate at the edge of chaos, likewise characterised by the network’s ability to maintain an optimal balance between rigidity (order) and randomness (chaos), where ANNs achieve maximum computational efficiency and robustness \cite{morales2021optimal}. 
Empirical studies suggest that ANNs trained and tuned to operate near this critical point exhibit superior performance, especially in dynamic and non-trivial environments\cite{feng2021optimal}, enabling them to optimise memory retention, responsiveness to inputs, and generalisation across new data which is particularly evident in models such as the recently developed Liquid Neural Networks (LNNs) \cite{hasani2020liquid}, where they have been shown to utilise continuous-time recurrent neural networks with dynamic parameters, allowing them to adapt more effectively to time-varying inputs and maintain criticality over extended periods. In contrast,  \cite{feng2021optimal} demonstrated that features learned in the chaotic phase that exceeds criticality, are not stable patterns, leading to a reduction in information processing capability, which renders the models unable to generalise and makes them fragile against input perturbations. This degradation in ANNs is analogous to biological organisms where deviations from criticality in the brain, either through its loss or excess, impair functionality and result in pathological conditions such as epilepsy and altered physiological states \cite{munoz2018colloquium}. 

However, whereas in nature neurobiological systems optimize their interactions with the environment by modifying the morphology of synaptic connections in response to past experiences \cite{Bianchi2020bio} and thereby continue to learn, ANNs are trained to operate at the edge of chaos with their weights generally remaining fixed thereafter, which limits their ability to incorporate new knowledge \cite{schmidgall2024brain}. Developing AI systems that can learn from their experiences and adapt to continuously changing environments is one of the primary challenges \cite{Bianchi2020bio}. While recent research has shown that transformer-based architectures can exhibit ongoing learning capabilities via in-context learning which operates via network activations rather than synaptic plasticity to solve novel tasks \cite{schmidgall2024brain}, other recent works \cite{Bianchi2020bio} have proposed ANN architectures that allow real-time parameter adjustments to facilitate continuous learning and adaptation. 
It is especially in these continuously adaptive systems that hold real potential for leaps in AI capabilities, that pose challenges for maintaining criticality where such adaptation may push the networks beyond the optimal critical point, leading to unstable behaviours that may not be easily detectable.

\subsection{AI Performance Benchmarks}

AI benchmarks are designed to test various performance capabilities in AI systems, and thus have the potential to detect the emergence of impaired functionality over time across different tasks. In the case of generative AI and specifically LLMs, AI benchmarks that go beyond testing natural language understanding and generation tend to include reasoning and logic benchmarks that measure problem-solving and multi-step and multi-hop reasoning abilities \cite{mcintosh2024inadequacies}. They also assess commonsense knowledge to evaluate understanding of everyday facts and relationships, as well as spatial and temporal reasoning, planning, and decision making \cite{mialon2023gaia}, and creativity and imagination. Other benchmarks evaluate contextual understanding and coherence, and adaptation to new tasks, probing the model's ability to quickly learn and adapt to new domains with minimal data, analogous to few-shot or zero-shot learning capabilities \cite{mcintosh2024inadequacies}. The peculiar and extraordinary attribute in the instance of LLMs is that these models are only trained to predict the next word/token in a sequence of text, and their sophisticated capabilities in reasoning and logic are considered emergent properties. These capabilities are therefore not explicitly taught but arise from the complex interplay of the entire network, indicating transfer learning that allows neural networks to perform well across a wide range of tasks by leveraging shared representations and patterns learned during pretraining. 

When performing AI benchmarking, these emergent capabilities are assessed separately to understand how well the model performs in each area. This functional decomposition does not imply that the model has distinct parts dedicated to each capability within its neural network, but rather that it can be evaluated for each skill individually. Achieving uniform optimisation in performance across multiple tasks is a challenge, analogous to maintaining and operating at the point of criticality across each distinct functional capability. Divergences from the optimal point of criticality may not be immediately apparent during model training and may only become detectable via extensive benchmark evaluations. While current transformer-based architectures aim to operate at the edge of chaos, future ANN architectures may incorporate adaptability and plasticity for continuous learning from experiences. Thus, both the current static ANNs and the potentially adaptable ANNs present vulnerabilities to exceeding criticality that are worth exploring.

\subsection{Literature summary}

The literature review underscores the rapid advancements in AI capabilities with the expectation of further significant developments in ANN architectures. A key insight is the importance of maintaining criticality and operating at the edge of chaos in ANNs to optimize their performance and stability. As AI systems become more complex and potentially capable of adapting and exhibiting neuronal plasticity akin to biological systems, managing criticality will become increasingly challenging. Research does not currently exist with a focus on modelling and detecting this hypothetical state of excess criticality. To that end, this study seeks to answer the following research questions:

\begin{enumerate}
    \item[(RQ1)]  Is it possible to construct a simulation framework that simulates AI systems becoming more complex over time, allowing a study of their performance beyond the point of criticality?
    \item[(RQ2)]  Is it possible to identify the emergence of instability in AI systems resulting from excessive criticality?

\end{enumerate}

\section{Theoretical Framework}

The proposed framework explores the dynamics governing the performance of AI systems, represented through a set of benchmarks, where we assume that each benchmark reflects different AI capabilities and by proxy, therefore the underlying complexity of an AI system. At the heart of this exploration is the study of possible behavioural patterns that may be observable in AI systems beyond a critical juncture where their accrued complexity precipitates a significant shift in operational dynamics, leading to unpredictable performance fluctuations. Under this assumption, the proposed framework will seek to both model hypothetical AI systems as they approach and exceed the criticality point, and will also attempt to detect if this has occurred. The list of symbols and their definitions for the subsequent section are shown in Table \ref{tab:definitions}.

\begin{table*}[htbp]
\centering
\caption{Table of symbols}
\begin{tabular}{rl}
\toprule
\textbf{Symbol} & \textbf{Description} \\
\midrule
\( n \) & Number of benchmarks modeled as agents within the system \\
\( P_i(t) \) & Performance of the \( i^{th} \) agent at time \( t \) \\
\( C(t) \) & System complexity at time \( t \), calculated as the weighted average performance \\
\( C_{\text{max}} \) & Critical threshold that triggers a shift to more volatile dynamics \\
\( \mu_{random} \) & Randomly generated mean for the normal distribution modelling performance gains \\
\( \sigma \) & Standard deviation for the normal distribution affecting performance gains \\
\( \text{excess\_complexity\_ratio} \) & Ratio of how much \( C(t) \) exceeds \( C_{\text{max}} \) \\
\( \text{volatility\_factor} \) & Factor that increases the variance in performance as \( C(t) \) exceeds \( C_{\text{max}} \) \\
\( \sigma_{\text{var}} \) & Standard deviation used to model increased performance variability under critical conditions \\
\( \mathbf{D} \) & Dataset representing performance metrics across simulations and time points \\
\( \theta^* \) & optimised threshold for detecting criticality \\
\( \mathcal{A} \) & Accuracy function used to determine the effectiveness of \( \theta^* \) \\
\( \tau \) & Hypothesized critical point in system dynamics for model calibration \\
\( f \) & Function that defines state transition of agents based on current state and environmental influences \\
\( \mathcal{E}_i \) & Environmental factors influencing the state transition of agent \( i \) \\
\( \Theta_i \) & Set of parameters defining the behaviour of agent \( i \) \\
\( \mathcal{N} \) & Normal distribution function used to simulate performance updates \\
\bottomrule
\end{tabular}\label{tab:definitions}
\end{table*}

\subsection{Formal Representation}

We introduce $n$ benchmarks (representing non-trivial capabilities), denoted by $P_i(t)$, representing the performance of a single hypothetical AI system on the $i^{th}$ benchmark at time $t$, where the performance ranges from $0 - 1.0$. The aggregate performance metric $C(t)$, reflecting the agent's overall capability and therefore its complexity, is computed as the weighted average\footnote{Certain AI competencies like planning and multi-hop reasoning can be associated with higher weightings to represent increased complexities.} of individual performances:

\begin{equation}\label{performance}
C(t) = \frac{1}{n} \sum_{i=1}^{n} w_i P_i(t),
\end{equation}
where $w_i$ denotes the weight assigned to AI agent's performance on the $i^{th}$ benchmark, reflecting its relative importance and contribution to the system's complexity allowing for different capabilities to be expressed as incurring varying levels of complexity.

\subsection{Dynamical behaviour at Critical Thresholds}

The system's evolution towards and beyond the critical threshold $C_{\text{max}}$ is captured through the dynamics of $C_i(t)$. This threshold signifies the point where operational dynamics transition from predictable improvements to stochastic variations due to increasing and unmanageable system complexity. While $ C(t) < C_{\text{max}}$, we model  an increase :

\begin{equation}
P_i(t+1) = \min\left(P_i(t) + \frac{\mathcal{N}(\mu_{random}, \sigma)}{1 + P_i(t)}, 1\right)
\end{equation}

\noindent where $\mathcal{N}(\mu_{random}, \sigma)$ denotes the normal distribution which models performance gains or decreases for the subsequent time-step and is governed by the randomly generated $\mu_{random}$ as the mean that represents the variability or uncertainty in the performance gain, and $\sigma$ as the standard deviation, initially set as 0.01.  If the criticality point is reached or exceeded, where \(C(t) > C_{\text{max}}\), then the effects of the phase transition to more volatile behaviour are modelled as follows:

\begin{equation}
\text{excess\_complexity\_ratio} = \max\left(0, \frac{C(t) - C_{\text{max}}}{C_{\text{max}}}\right)
\end{equation}

\begin{equation}
\text{volatility\_factor} = \exp\left(1 + \text{excess\_complexity\_ratio}\right)
\end{equation}

\begin{equation}
P_i(t+1) = P_i(t) + \mathcal{N}  \times \text{volatility\_factor} \times \sigma_{\text{var}} 
\end{equation}

\noindent where \(\mathcal{N}\) generates a sample from a standard normal distribution, and \(\text{volatility\_factor}\) is an exponential factor increasing the variance in response to system complexity which is also amplified by \(\sigma_{\text{var}}\) that acts as an additional stochastic component for scaling the performance under critical conditions.

To ensure that \(P_i(t+1)\) remains within the bounds [0, 1], the following condition is applied:

\begin{equation}
P_i(t+1) = \max(\min(P_i(t+1), 1), 0)
\end{equation}

\noindent This adjustment handles scenarios where the calculated performance exceeds the upper limit by clamping the value within the specified range, thus preventing any unrealistic performance metrics.

\subsection{Criticality Detection}

Once criticality has been exceeded, detecting the onset of this phase transition into a less stable and more unpredictable state becomes practically useful from an application perspective. The proposed detection approach involves analysing $\mathbf{D} \in \mathbb{R}^{m \times n}$, which represents data on the performance metrics $C(t)$ across $n$ simulations over $m$ time points. Since each simulation reaches a predefined criticality point at a different time, all data $\mathbf{D} \in \mathbb{R}^{m \times n}$ is first aligned with respect to the critical point, and subsequently we employ a derivative-based analysis to monitor the rate of change in the standard deviation (SD) of these metrics (visually depicted in Figures \ref{fig:derivativedetection}(a-d)), which is explored for its potential to be indicative of critical transitions. We define $S(t)$ as the SD of the performance metrics at time $t$ for each benchmark across all simulations. The derivative of $S(t)$, denoted as $S'(t)$ therefore captures the dynamics of performance variability:

\begin{equation}
S'(t) = \frac{dS(t)}{dt},
\end{equation}

Each simulation provides a unique trajectory of $S'(t)$, reflecting individual variations in response to evolving system conditions. The criticality point is determined by observing when the derivatives exceed a threshold:

\begin{equation}
\theta^* = \underset{\theta}{\mathrm{argmax}} \; \mathcal{A}(\theta; \mathbf{D}, \tau),
\end{equation}

Here, $\mathcal{A}(\theta; \mathbf{D}, \tau)$ measures the accuracy of detecting criticality for a given threshold $\theta$, where $\tau$ is the hypothesised critical point in the system dynamics. The threshold $\theta^*$ is empirically determined from a range of simulations to ensure it optimally balances sensitivity to changes and specificity in detecting criticality and can be evaluated on separate hold-out simulation datasets for generalisability.

\subsection{Optimisation of Detection Threshold}

To optimise the detection threshold $\theta^*$, the framework uses Stochastic Gradient Descent (SGD) by iteratively adjusting it in response to the gradients of a loss function calculated from the criticality detection accuracy:

\begin{equation}
\theta_{i+1} = \theta_i - \eta \nabla \mathcal{L}(\theta_i; \mathbf{D}, \tau),
\end{equation}

where $\eta$ represents the learning rate and $\nabla \mathcal{L}(\theta; \mathbf{D}, \tau)$ is the gradient of the loss function, reflecting discrepancies in criticality detection accuracy. This process continues until the difference between successive threshold values falls below a defined tolerance $\epsilon$, indicating convergence to an optimal threshold:

\begin{equation}
|\theta_{i+1} - \theta_i| < \epsilon.
\end{equation}

\subsection{Agent-Based Modelling}

ABM is employed to simulate emergent behaviours within the hypothetical AI systems by conceptualising each performance benchmark as an autonomous agent \(a_i\), each encapsulating a distinct aspect of the system's capabilities. These agents are modelled to interact within a dynamic environment, collectively influencing the system's behaviour as it approaches or surpasses critical complexity thresholds.

\subsubsection{Agent State Transition}

Each agent's state, \(P_i(t)\), evolves based on a combination of internal decision-making processes and interactions with its environment:

\begin{equation}
P_i(t+1) = f(P_i(t), \mathcal{E}_i(P_i(t)), \Theta_i),
\end{equation}

where \(f\) denotes a function that integrates the current state \(P_i(t)\), the environmental influences \(\mathcal{E}_i(P_i(t))\), and a set of parameters \(\Theta_i\) that govern the agent's behaviour. Here, \(\Theta_i\) encompasses decision thresholds, performance adjustments and adaptability levels, and other behavioural rules. The environmental influences \(\mathcal{E}_i\) include both direct interactions with other agents (benchmark performances) and external systemic conditions, reflecting how interactions at the micro-level (individual benchmarks) can affect macro-level system dynamics (overall system performance).

\subsubsection{Inter-Agent Dynamics}

The interaction dynamics among agents lead to emergent behaviours in the system, particularly as it relates to its stability and response to critical thresholds being reached and the extent to which they are exceeded:

\begin{equation}
I_{a_i}(t) = g(\{P_{a_j}(t)\}_{j \in \mathcal{N}_{a_i}}),
\end{equation}

where \(I_{a_i}(t)\) encapsulates the cumulative effects of interactions on agent \(a_i\) from its neighbours \( \mathcal{N}_{a_i} \), and \(g\) is a function that models these inter-agent influences. This model captures how the aggregate interactions and performances of individual benchmarks contribute to emergent system properties, such as increased instability or performance degradation, that are not predictable by observing single agents in isolation.

\subsubsection{Emergent Behaviour and Criticality}

The emergent behaviours observed in the modelled system, including transitions to critical states, arise from the interplay of the interactions captured by the aggregate performance metric \(C(t)\) (Equation \ref{performance}), which is indicative of the overall system complexity and serves as an indicator of the proximity to critical thresholds. As system complexity increases, due to compounded interactions and the accumulation of minor changes in benchmark performances/behaviours, the system may exhibit sudden shifts in overall performance, manifesting excess criticality as an emergent property.


\section{Methodology}

\subsection{Simulation Framework and Dynamics}

We conducted 200 simulations across various system scales to explore AI system dynamics near and beyond critical complexity thresholds. We used 2, 5, 10, and 20 benchmarks represented as agents within the Python-based Mesa\cite{python-mesa-2020} framework. This agent-based modelling approach allowed for a detailed analysis of hypothetical emergent behaviours under different levels of complexity and stress with respect to the stated assumptions.

\subsubsection{Agent Initialisation and Performance Dynamics}
Each agent simulated a performance benchmark and was initialised with a performance level \( P_i(0) \) randomly assigned between 0 and 0.7 to mirror the diverse capabilities seen in real AI systems, and each agent was assigned uniform weights. The performance evolution of agents was modelled through:
\begin{equation}
P_i(t+1) = P_i(t) + \frac{\mathcal{N}(\mu_{gain}, 0.01)}{1 + P_i(t)},
\end{equation}
where \( \mu_{gain} \) randomly varied within [0, 0.05] bounds to reflect stochastic learning dynamics. In our simulation design, the critical threshold \( C_{\text{max}} \) was set at 0.8 across all experimental setups. This threshold was not empirically derived but was instead an arbitrary marker used to trigger the onset of systemic volatility and to test the system's behaviour under increased operational complexity. Post-critical threshold \( C_{\text{max}} = 0.8 \), the performance updates for the agents incorporated increased volatility:
\begin{equation}
P_i(t+1) = P_i(t) + \mathcal{N}(0, \sigma_{\text{var}}) \times \text{volatility\_factor}_i,
\end{equation}
modelling the heightened unpredictability as complexity exceeded the predefined threshold, where \( \mathcal{N}(0, \sigma_{\text{var}}) \) represents a random noise with mean 0 and standard deviation \( \sigma_{\text{var}} \), reflecting the system’s current state relative to \( C_{\text{max}} \). The \( \text{volatility\_factor}_i \) was randomly predefined for each agent within the [0, 1.0] bounds.

\subsection{Criticality Detection and Optimisation}

Critical thresholds trigger systemic volatility, challenging the system's stability and performance. Criticality is algorithmically detected by assessing performance variability with thresholds optimised through SGD:
\begin{equation}
\theta_{\text{new}} = \theta - \eta \times \frac{\text{Loss}(\theta + \epsilon) - \text{Loss}(\theta)}{\epsilon},
\end{equation}

where the learning rate (\(\eta\)) was initially set to \(1 \times 10^{-5}\) and a tolerance level for convergence at \(1 \times 10^{-6}\) while \(\epsilon\), used for gradient estimation was set at \(1 \times 10^{-4}\). Each of these values was empirically derived and validated against additional simulation data for reproducibility in detecting critical transitions. Algorithm \ref{alg} describes the entire criticality threshold procedure. 

\begin{algorithm*}
\caption{Stochastic Gradient Descent for Threshold Optimisation}
\fontsize{9pt}{10pt}\selectfont
\begin{algorithmic}[1]
\Require
\Statex $D$: Dataset containing performance values across simulations with time points.
\Statex $actual\_critical\_point$: The critical time point from simulation data to which all simulations have been aligned.
\Statex $initial\_threshold$: Initial guess for the threshold value.
\Statex $learning\_rate = 1 \times 10^{-5}$: Step size for each iteration's threshold update.
\Statex $tolerance = 1 \times 10^{-6}$: Convergence criterion for the optimisation.
\Statex $max\_iterations = 1000$: Maximum number of iterations.
\Statex $\Call{DetectedCriticalTimePointsBasedOnThreshold}{D, threshold}$: Function returns points in time when criticality is detected wrt. the threshold.

\Ensure
\Statex The optimised threshold value that best identifies the critical point.

\State $threshold \gets initial\_threshold$
\For{$iteration \gets 1$ \textbf{to} $max\_iterations$}
    \State $gradient \gets \Call{EstimateGradient}{D, actual\_critical\_point, threshold}$
    \State $new\_threshold \gets threshold - learning\_rate \times gradient$

    \If{$|new\_threshold - threshold| < tolerance$}
        \State \textbf{break}
    \EndIf
    \State $threshold \gets new\_threshold$
\EndFor
\State \textbf{return} $threshold$

\Function{EstimateGradient}{$D$, $actual\_critical\_point$, $threshold$, $epsilon = 1 \times 10^{-4}$}
    \State $loss\_at\_threshold \gets \Call{LossFunction}{D, actual\_critical\_point, threshold}$
    \State $loss\_at\_threshold\_plus\_epsilon \gets \Call{LossFunction}{D, actual\_critical\_point, threshold + \epsilon}$
    \State $gradient \gets (loss\_at\_threshold\_plus\_epsilon - loss\_at\_threshold) / \epsilon$
    \State \textbf{return} $gradient$
\EndFunction

\Function{LossFunction}{$D$, $actual\_critical\_point$, $threshold$}
    \State $critical\_points \gets \Call{DetectedCriticalTimePointsBasedOnThreshold}{D, threshold}$
    \State $accuracy \gets$ \% correct wrt. threshold $actual\_critical\_point <= critical\_points <= actual\_critical\_point + 10))$
    \State \textbf{return} $-accuracy$
\EndFunction
\end{algorithmic}\label{alg}
\end{algorithm*}

\subsection{Model Testing and Validation}

The criticality detection thresholds were derived for each agent size configuration. Subsequently, the thresholds were validated and tested via cross-validation against 100 additional simulations for each of the four agent configurations to establish the robustness of the criticality detection mechanism\footnote{All simulation data and code are available on the project's GitHub repository https://github.com/teosusnjak/AGI-and-criticality}. A simple measure of accuracy in the form of the percentage of criticality events detected within a predefined time window was used. The time window denoting a correct detection was defined as 10 time steps post actual criticality which is depicted in Figures \ref{fig:derivativedetectionhist}(a-d). Accuracies for training and test simulation data were repeated 20 times each and reported as a mean, together with the standard deviations. 

\section{Results}

The results section is presented in two parts, where the first presents the trends and dynamics of the modelled systems at the point of criticality and beyond; while the second part examines how the onset of excess criticality can hypothetically be detected in real-world scenarios. 

\subsection{System dynamics at criticality}

\begin{figure}[hbt]
\centering
\subfloat[2 benchmarks.]{\includegraphics[width=0.45\textwidth]{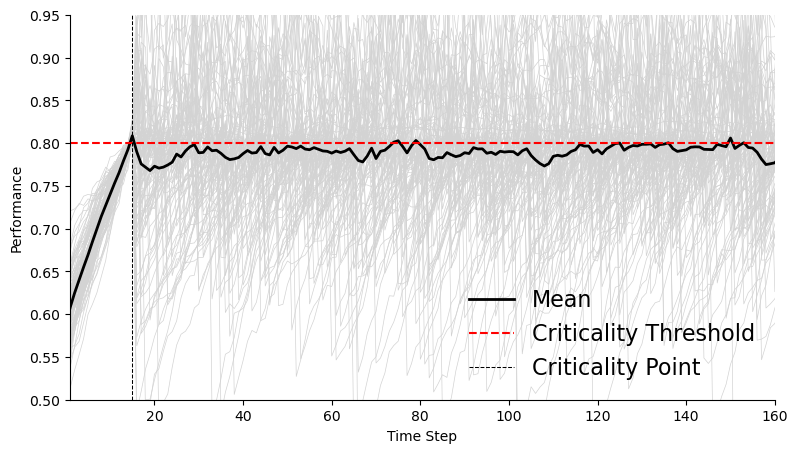}} 
\subfloat[5 benchmarks]{\includegraphics[width=0.45\textwidth]{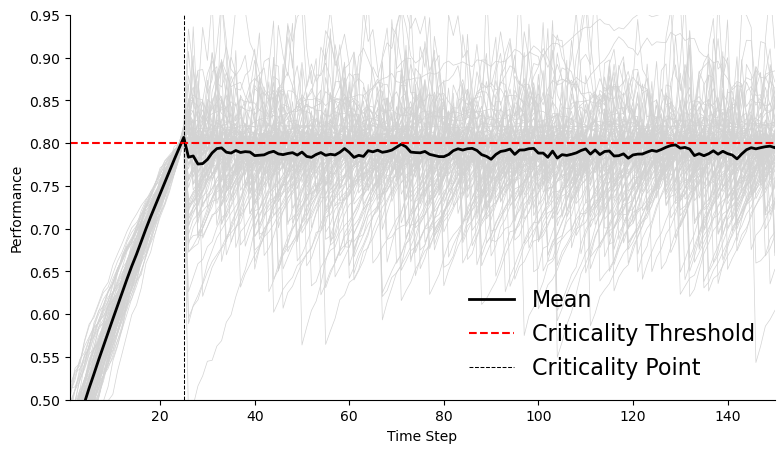}} \\
\subfloat[10 benchmarks]{\includegraphics[width=0.45\textwidth]{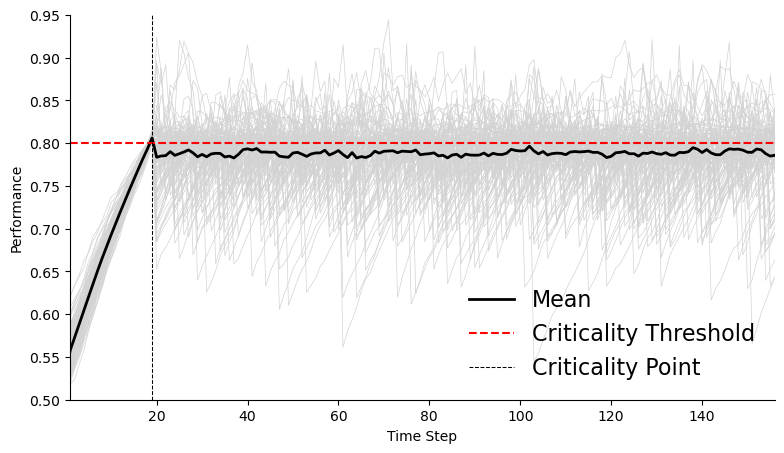}} 
\subfloat[20 benchmarks]{\includegraphics[width=0.45\textwidth]{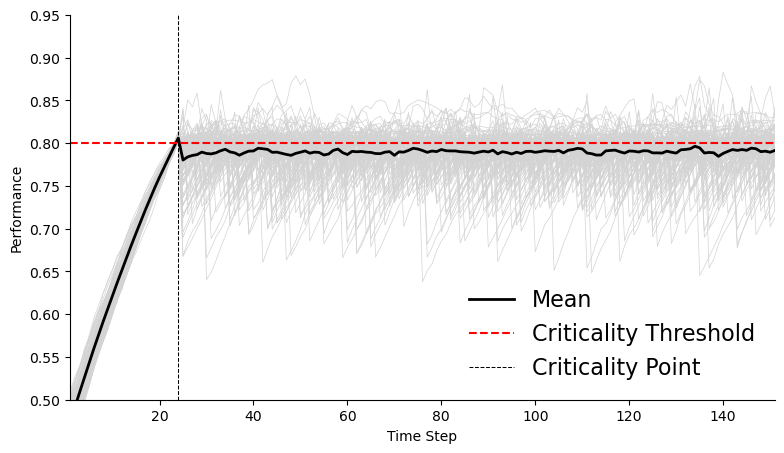}} \\
\caption{Aggregate performances across all simulations, indicating the criticality point to which all simulations have been aligned and an predetermined criticality threshold.}
\label{fig:means}
\end{figure}

The convergence of AI systems toward the threshold of criticality, depicted in Figure \ref{fig:means}(a-d), illustrates the mean performances over time for varying benchmark sizes across 100 simulations. These simulations have been synchronised at the criticality point, providing a coherent baseline from which to observe the systems' behaviour. A clear demarcation is indicated by the criticality threshold as a horizontal red dashed line, with the criticality point marked by a vertical line. Notably, the transition into criticality is marked by increased fluctuations in performance, suggesting the emergence of complex system dynamics. The AI systems with fewer benchmarks demonstrate pronounced instability after reaching and exceeding criticality, as seen in the widening spread of the grey lines, reflecting a transition from predictable growth to a stochastic regime. For systems with a larger number of benchmarks, there is a notable decrease in the performance variance as the number of benchmarks increases.  
The mean performance line, depicted in black, remains relatively steady before the criticality threshold, indicating ordered growth. However, the increase in performance variability post-criticality suggests that this ordered state gives way to more erratic behaviour, where the system's performance is influenced by complex interactions and inherent noise within the AI system.

\begin{figure*}[hbt]
\centering
\subfloat[2 benchmarks.]{\includegraphics[width=0.45\textwidth]{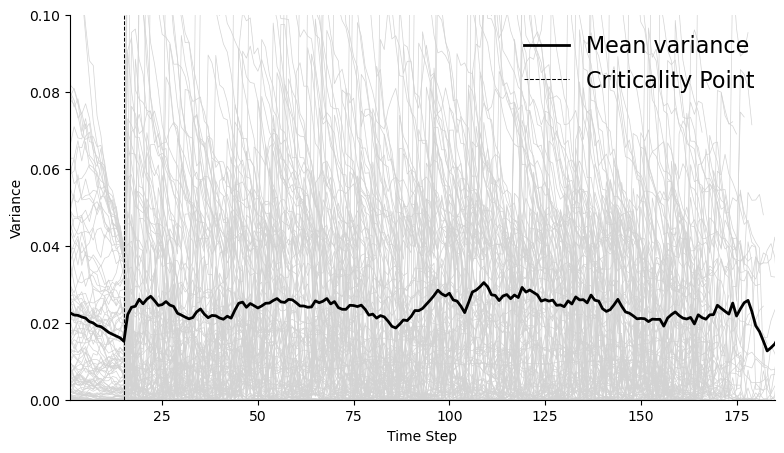}} 
\subfloat[5 benchmarks.]{\includegraphics[width=0.45\textwidth]{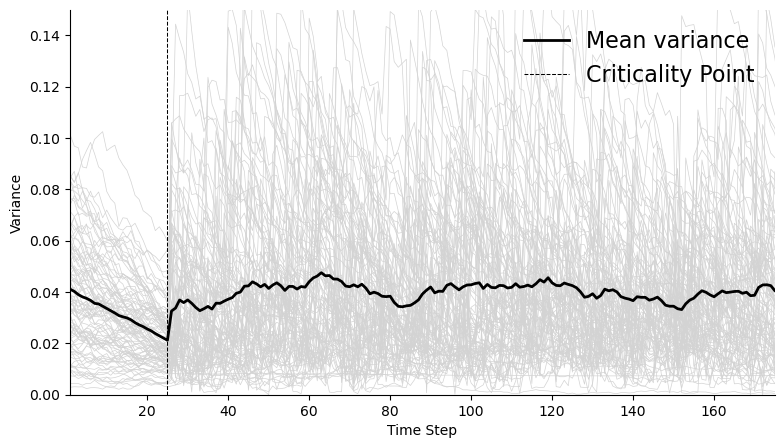}} \\
\subfloat[10 benchmarks.]{\includegraphics[width=0.45\textwidth]{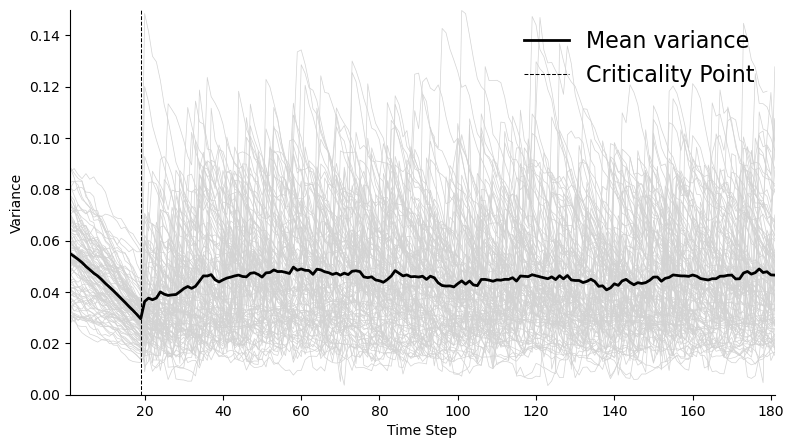}} 
\subfloat[20 benchmarks.]{\includegraphics[width=0.45\textwidth]{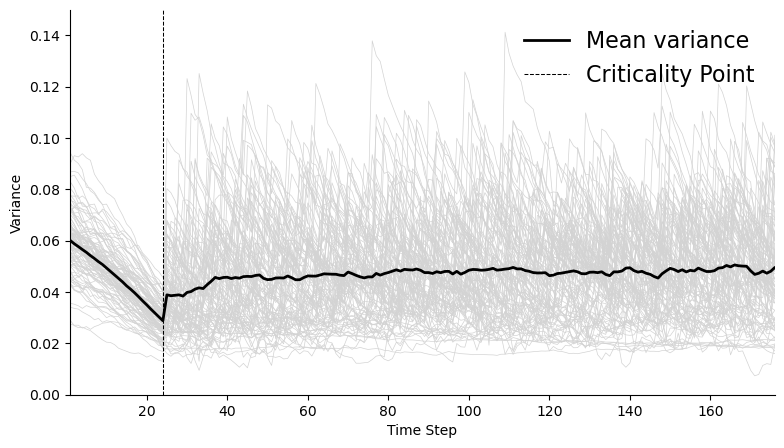}} \\
\caption{Variance of all performances across all simulations, indicating the criticality point to which all simulations have been aligned and a predetermined criticality threshold.}
\label{fig:variance}
\end{figure*}

The subsequent plots in Figure \ref{fig:variance}(a-d) show the variance of the performances of all the benchmarks over time across all 100 simulations as the AI systems approach the criticality point and exceed the criticality threshold. The mean of all the variances is also depicted in order to highlight trends and inflexion points. The figures indicate the changes in the fingerprint of the variance which can hypothetically be leveraged for detection purposes of criticality in a real-world scenario. The figures indicate that as expected, the variance and instability are notably higher for modelled systems with fewer benchmarks, while the variance reduces while the observability of a clear inflexion point increases as more benchmarks are deployed.

\subsection{Detecting criticality}

\begin{figure*}[htb]
\centering
\subfloat[2 benchmarks.]{\includegraphics[width=0.45\textwidth]{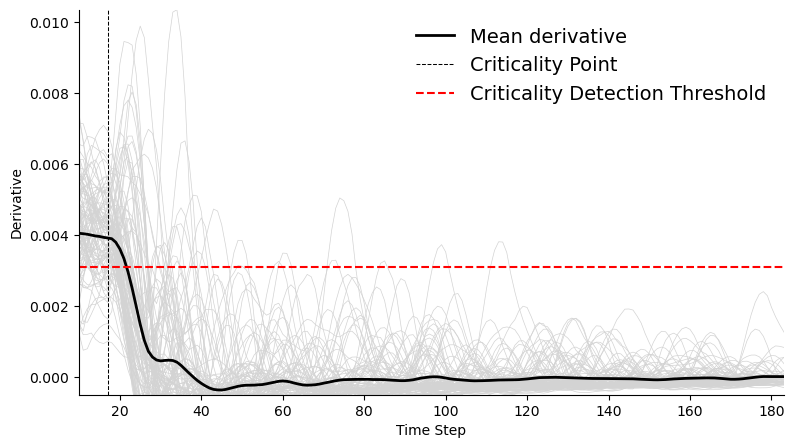}} 
\subfloat[5 benchmarks.]{\includegraphics[width=0.45\textwidth]{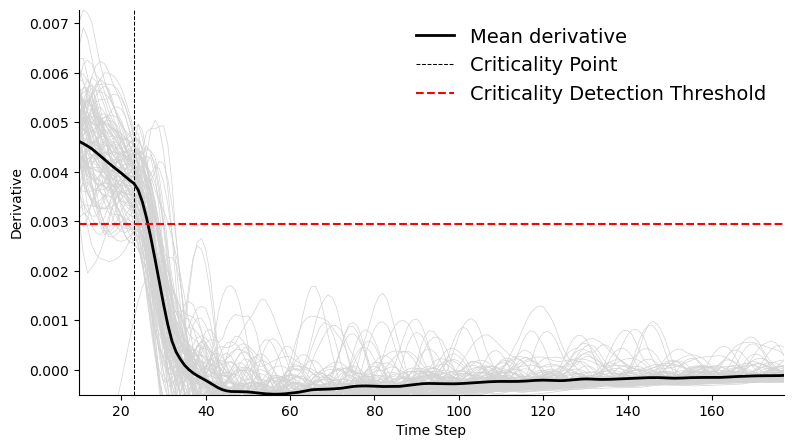}} \\
\subfloat[10 benchmarks.]{\includegraphics[width=0.45\textwidth]{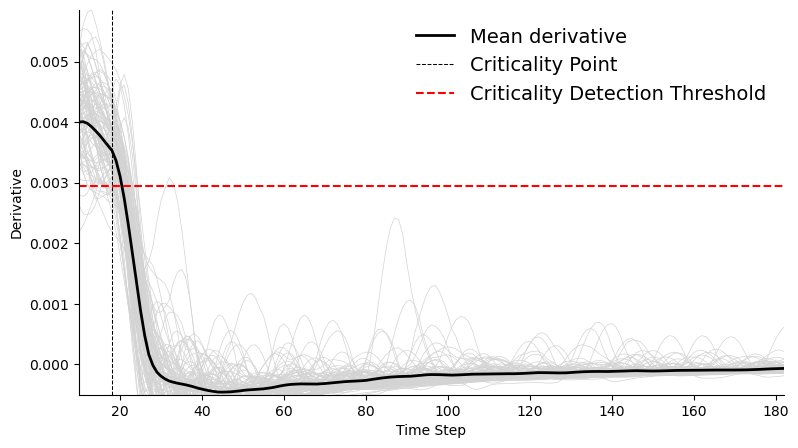}} 
\subfloat[20 benchmarks.]{\includegraphics[width=0.45\textwidth]{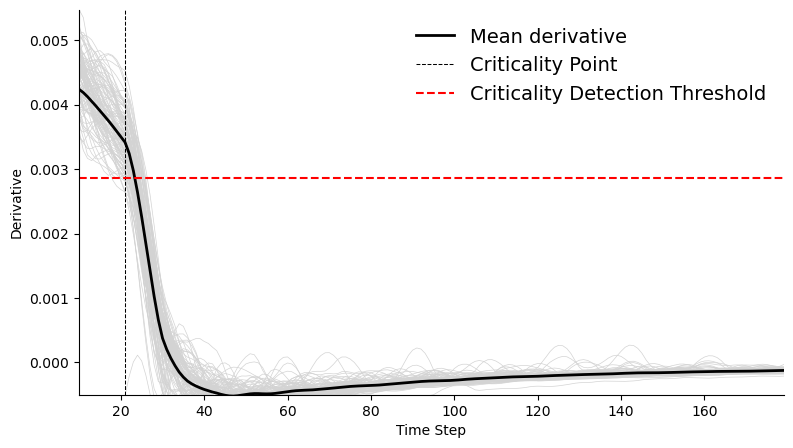}} \\
\caption{Variance of all performances across all simulations, indicating the criticality point to which all simulations have been aligned and a predetermined criticality threshold.}
\label{fig:derivativedetection}
\end{figure*}

The following sequence of plots in Figure \ref{fig:derivativedetection}(a-d) depict test dataset results from a new set of 100 simulations. The figures show the derivative value calculated over the standard deviations of all benchmark performances expressed as an expanding window from the initial time point. The mean of all derivatives is depicted to highlight the aggregate trend. The figures also depict both the aligned critical point for all the next simulation experiments as well as the derived threshold calculated from the previous training dataset of simulations for determining the optimal decision threshold. This threshold derived from the training dataset would then be evaluated for identifying criticality based on the derivative of the standard deviations of performances.

All figures indicate that immediately following the entry into the criticality phase, a pronounced change in the slope of the derivative becomes observable in most cases. The figures also show that the optimal threshold derived for detecting criticality from the training data visually corresponds well with the test data. Since detecting a change in the slope of the instability in the performances requires several time steps to transpire and will vary to some degree between different simulations, it remains to be demonstrated how soon post-criticality, this transition can be detected. For the purposes of these experiments, we selected any detection that falls immediately within 10-time steps post-criticality to be a correct positive detection. Based on this criterion, Table \ref{table:benchmark_results} shows the percentage of correct classifications that were achieved on both the training and test datasets across all benchmark sizes. It is evident that as the number of benchmarks increases, the detection of variability becomes more reliable and the difference between the training and test errors also decreases.

\begin{table}[t!]
\centering
\caption{The percentage of correct criticality detections with standard deviations on the training dataset as well as correct classifications on test dataset results across 20 repetitions of all the simulations.}
\begin{tabular}{rll}
\toprule
\textbf{Benchmarks} & \textbf{Training Dataset} & \textbf{Test Dataset} \\
\midrule
2 & 61.8 $\pm$ 14.5 & 62.8 $\pm$ 5.6 \\
5 & 87.7 $\pm$ 5.5  & 86.4 $\pm$ 3.7 \\
10 & 92.5 $\pm$ 5.5 & 88.2 $\pm$ 2.4 \\
20 & 96.4 $\pm$ 2.1 & 95.5 $\pm$ 1.7 \\
\bottomrule
\end{tabular}

\label{table:benchmark_results}
\end{table}

\begin{figure*}[htb]
\centering
\subfloat[2 benchmarks.]{\includegraphics[width=0.45\textwidth]{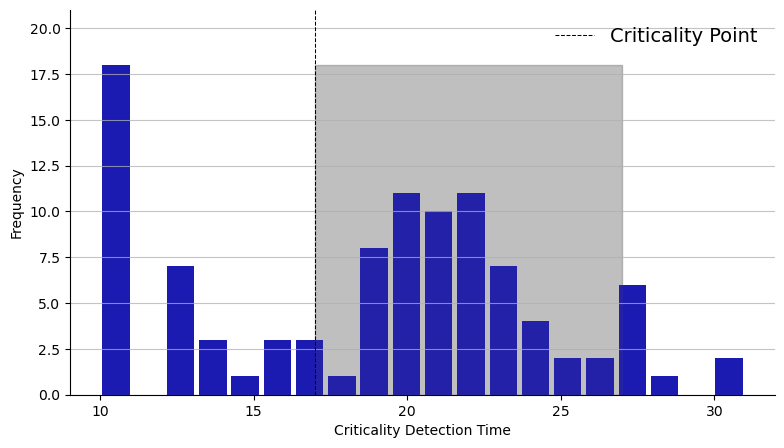}} 
\subfloat[5 benchmarks.]{\includegraphics[width=0.45\textwidth]{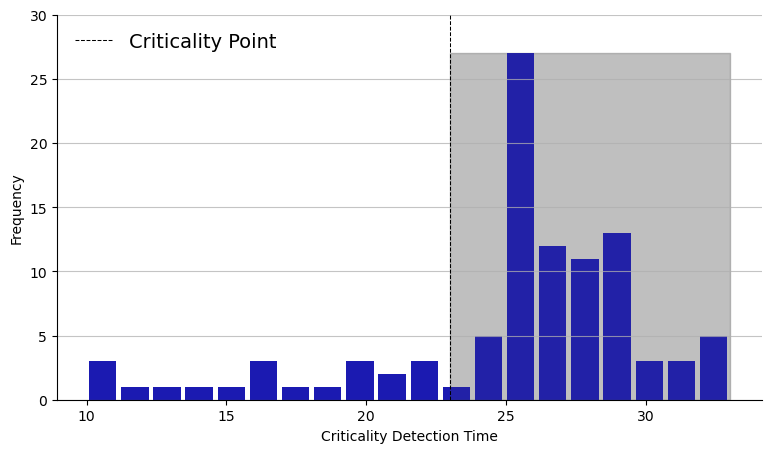}} \\
\subfloat[10 benchmarks.]{\includegraphics[width=0.45\textwidth]{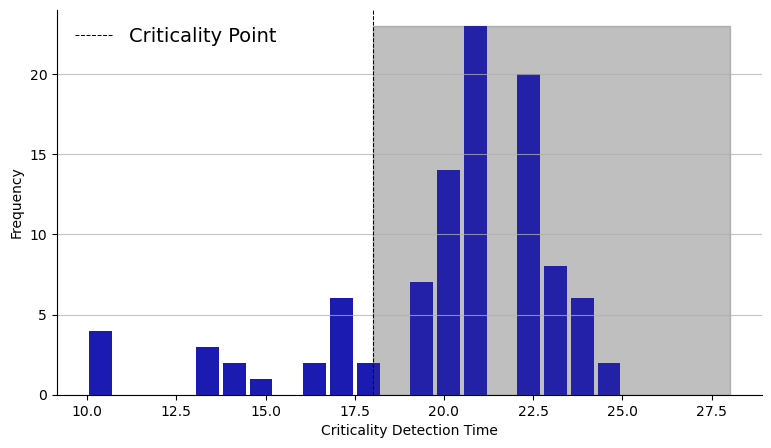}} 
\subfloat[20 benchmarks.]{\includegraphics[width=0.45\textwidth]{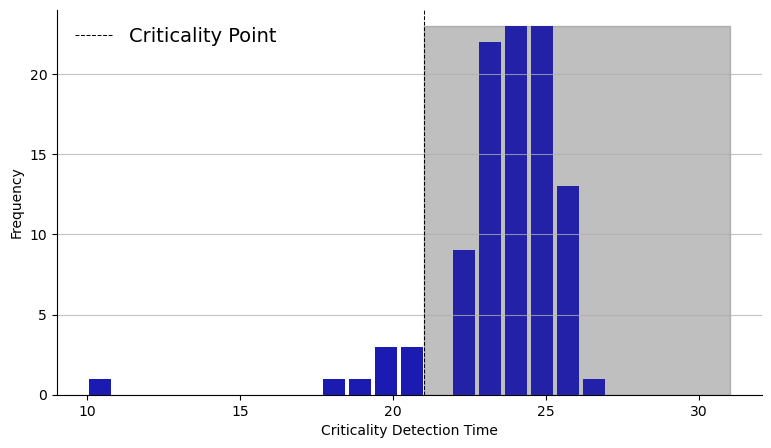}} \\
\caption{Histograms depicting the distribution of time points at which criticality was detected on the test dataset with the shaded area indicating the beginning of the actual criticality and the end of the detection window period denoting correct classifications.}
\label{fig:derivativedetectionhist}
\end{figure*}

The histograms in Figure  \ref{fig:derivativedetectionhist}  depict the distribution of the criticality detection times in a new set of 100 simulations, represented as test datasets. These simulations utilise the threshold derived from the training dataset to determine the optimal decision threshold to detect criticality. The shaded region in each subplot illustrates the window of time considered for accurate detection, extending 10 steps beyond the identified criticality point. It is evident from the histograms that most of the detections occur within this acceptable window, signifying a reliable threshold.
As the number of benchmarks increases from 2 to 20, there is a noticeable improvement in the accuracy of criticality detection, with the distribution of detection times becoming increasingly concentrated within the defined window. This observation underscores the value of relying on a diverse set of benchmarks in enhancing the ability to detect and adapt to the onset of criticality. The consistent alignment of the derived threshold with the criticality point across all test datasets confirms the robustness of the approach used to establish this threshold. These results suggest that the method used for criticality detection is effective and could potentially be applied to real-world scenarios where early identification of critical states is crucial.

\section{Discussion}

This paper demonstrates an approach to constructing a simulation framework modeling AI systems becoming more complex over time, enabling a study of their possible performance patterns beyond criticality (RQ1).  The utility of this framework relies on several key assumptions, such as the direct relationship between benchmark performance and system complexity, and the nature of criticality in AI systems. Given the assumptions, this paper also shows that it is possible to detect the emergence of instability in AI systems resulting from excessive criticality through a proposed method that leverages patters in the rate of change in performance metrics of AI systems across various benchmarks (RQ2).
The next logical question is whether the newest versions of transformer-based AI systems, like large language models (LLMs), despite their growing capabilities and remarkable performance, represent true growth in intelligence with corresponding complexity that could result in criticality. Are they likely to keep increasing in complexity past a critical limit post-training, or will their current performance plateau within the limitations of present artificial neural network (ANN) architectures?

\subsection{Lack of real reasoning, planning and world model constructions}
Recent studies \cite{mialon2023gaia,mialon2023augmented} demonstrate that current LLM models struggle with complex reasoning tasks that humans can perform relatively easily and fail to display a deep understanding, relying instead on surface-level pattern recognition or exploiting dataset biases. 
This observation stems from the inherent architectural constraint where the computational effort for generating responses is not dynamically allocated based on the complexity of the task but is rather fixed per token generated. This mechanism starkly contrasts with human cognitive effort, which scales with task complexity. This phenomenon can be linked to the dichotomy between System One and System Two-type thinking \cite{kahneman2011thinking} which provides a lens for understanding the disparity between human reasoning and the current capabilities of LLMs. System One encompasses fast, automatic, and often subconscious thought processes, akin to what LLMs might simulate through immediate, pattern-based responses, while System Two, involves slower, more deliberate, and effortful reasoning processes, requiring cognitive resources that LLMs are architecturally not designed to emulate. Human reasoning involves an interplay between these systems, where System Two can override or modulate responses generated by System One, meanwhile, this adaptive allocation of cognitive resources is a sign of increasing degrees of true complexity via the ability to engage in deep, reflective thought is absent in LLMs as they currently stand. 
A key feature of a genuine ability to reason over non-trivial tasks is to exhibit planning capability. 
Again, recent works \cite{lecun2022path,mialon2023augmented} have also contended that current LLMs are restricted to System One tasks that are intuitive and reflexive, rather than tasks that require logic and deliberate analysis involved the System Two level. Others \cite{Kambhampati2024can} have categorically claimed that LLMs cannot reason nor plan, but can only give an impression of this capability. Meanwhile, other recent works \cite{west2023generative,oh2024generative} have even more strongly contended that the impressive generative ability of LLM-based agents may not at all be a reflection of their understanding capability but is merely a function word prediction. Indeed, current AI systems lack a crucial aspect of human intelligence which includes rich internal models of the world from which, causal inferences, simulations and predictions of events can be made, together with both reasoning and planning in new situations and a continuous update of our knowledge based on experiences \cite{Mitchell2023challenge}. It remains a subject of debate if LLMs can effectively learn an internal world model from ungrounded form alone as found in text data, without direct sensory input or interaction with the physical world \cite{wu2024visualizationofthought}. For these reasons, it seems more probable that current LLM-based AI systems will likely plateau in their performance rather than accrue excess complexity that crosses beyond criticality.

\subsection{Inefficiencies of inductive data-driven learning}
The limitations inherent in the simplistic objective functions used to train transformer-based LLMs suggest that we are not yet engaging with truly complex systems capable of deep world perception and understanding \cite{Mitchell2023challenge}. 
Human-level intelligence involves a dynamic interplay of cognitive processes, including abstract reasoning, problem-solving, understanding causal relationships and the ability to learn from minimal data. The objective functions used by LLMs, designed to minimise a narrowly focused loss, inadvertently promote shortcut learning, where LLMs learn to exploit statistical regularities and dataset biases to achieve high performance on tasks without a genuine grasp of the underlying phenomena \cite{Mitchell2023how}. This form of learning which is characterised by taking the path of least resistance towards loss minimisation, leads to models that do the right things for the wrong reasons, revealing a superficial understanding while maintaining a substantive capability. Studies \cite{lightman2023lets} show that within the domain of logical reasoning, trained models regularly use incorrect reasoning to reach the correct final answers when trained to predict the right outcome. Furthermore, the necessity of prompt engineering in LLMs which is a practice absent in human communication, underscores the artificiality and limitations of these models. Unlike humans who comprehend and interact with the world through a rich tapestry of experiences and cognitive processes, LLMs currently rely on carefully crafted prompts to access and navigate their knowledge space with large variability in results from small differences in prompts which further suggests their lack of true intelligence and underlying complexity. In short, the current transformer-based AI training paradigms are essentially data-driven where they focus on pattern recognition and prediction rather than on real understanding \cite{Mitchell2023challenge}.

There also exist other compelling reasons suggesting that while current LLMs (and their multi-modal AI extensions) represent a profound and undeniably significant leap forward with respect to AI capabilities, there are limiting factors that may result in current LLM agents eventually converging to performance plateaus. Firstly, LLMs are primarily trained on linguistic data, which itself is an artefact of human intelligence, reflecting only one dimension and thus a subset of human cognitive capabilities which is unable to encapsulate the entirety of human intellect or the complexities of real-world phenomena. Therefore, achieving human-like intelligence solely through this modality, in theory raises doubts \cite{Mitchell2023challenge} especially regarding the achievement of near-AGI or beyond to superintelligence. Our linguistic data resources are also finite. High-quality linguistic datasets necessary for extracting increased performances from LLMs are also decreasing since they have likely already been deployed for current model training purposes. Another important recent finding  \cite{udandarao2024zeroshot} examining the feasibility of such data-intensive inductive approaches to increasing capabilities in LLMs has revealed that model performance in understanding concepts improves linearly only when their frequency in the training data increases exponentially. This finding highlights potentially fatal inefficiencies in the data utilisation of LLMs trained by transformer-based architectures since the requirement of providing exponentially more data to achieve linear performance gains is not sustainable. 

\subsection*{The need for robust benchmarks}
Notwithstanding the current limitations of LLMs and indications of their architecture-defined complexity ceiling, there is a need to evolve the benchmarks for measuring their genuine intelligence and underlying complexity \cite{Mitchell2023how}. To truly gauge the sophistication of AI systems and their drift towards and beyond criticality, benchmarks need to become more diverse and robust, encapsulating the multi-dimensional facets of intelligence such as deep reasoning, planning, and conceptualisation capabilities. Present benchmarks fail to capture the comprehensive nature of intelligence, focusing instead on narrow domains that do not adequately reflect the emergent complexities of the prevailing AI systems \cite{mialon2023augmented}. Overall, evaluating new AI systems will require researchers to rethink benchmarks since static benchmarks are regarded as broken benchmarks in the making \cite{mialon2023gaia} due to data contamination and gaming of performances. What is needed, is an evolution and continuous updates of the benchmarks year-by-year by removing, adding and modifying questions and the addition of new ones to better assess the generalisation and robustness of AI systems  \cite{mialon2023gaia}.

\subsection*{Practical implications, study limitations and future work}

From a practical standpoint, the approaches outlined in this study can be implemented with immediate effect on the existing benchmarks and a set of criticality decision thresholds can be empirically derived and monitored over time. In terms of study limitations, while our investigation into AI system dynamics, particularly through the prism of LLMs and complexity theory has unearthed insights into potential criticality emergence, it is nonetheless constrained by the current benchmarking methodologies and the simplifications in our modelling approach. Current benchmarks do not fully capture the multidimensional and evolving complexity of AI systems which necessitates the development of holistic, robust benchmarks that integrate complexity measures to accurately reflect AI's capabilities. Moreover, the portrayal of AI capabilities as independent agents oversimplifies the intertwined nature of these systems where capabilities often overlap, leading to emergent behaviours that our current model, with its linear stochastic elements, may not adequately predict.

To advance our understanding and prediction of AI system behaviours especially near critical points, future research should innovate in benchmark design that captures the broader spectrum of AI capabilities and system complexities. This should involve not just expanding the range of reasoning tasks but eventually also require measuring the adaptability and the ability of ANNs to integrate new knowledge and experiences. Additionally, future criticality modelling efforts ought to evolve to reflect the non-linear, interdependent dynamics within AI systems.

\section{Conclusion}

Our study investigated the hypothetical constraints of AI systems through the lens of complexity theory and criticality. We leveraged agent-based modelling (ABM) combined with a set of well-defined assumptions, we identified scenarios where increasing complexity, represented by performance capabilities, may push AI systems to a critical threshold. Beyond this critical point, performance shifts from predictable improvements to erratic and unstable behaviour. Practically, we developed a detection method for identifying criticality in AI systems, showcasing its potential applicability in real-world scenarios with a relevance to large language models. We also conducted an extensive review of recent literature to underpin our findings, drawing from a wide array of empirical and theoretical studies that supported our conclusions.
Ultimately, our findings advocate for measured optimism regarding AI growth within the context of current transformer-based architectures and a very cautious outlook on the prospects of current data-driven deep learning systems achieving AGI. Future research should extend the proposed methodologies for modelling criticality within AI systems, reflecting non-linear dynamics, together with the creation of a broader array of benchmarks that ensure a more robust and comprehensive evaluation of the capabilities of emerging AI technologies.

\bibliographystyle{unsrtnat}

\end{document}